\documentclass{article} 
\usepackage{iclr2027_conference,times}

\usepackage{amsmath,amsfonts,bm}

\def\eqref#1{equation~\ref{#1}}

\def\1{\bm{1}}

\DeclareMathAlphabet{\mathsfit}{\encodingdefault}{\sfdefault}{m}{sl}
\SetMathAlphabet{\mathsfit}{bold}{\encodingdefault}{\sfdefault}{bx}{n}

\usepackage{hyperref}
\usepackage{url}
\usepackage{graphicx}
\usepackage{booktabs}
\usepackage{array}
\usepackage{multirow}
\usepackage{placeins}
\usepackage{enumitem}

\title{\raggedright PlayGen2: Learning Real-Time\\
Playable Worlds with Explicit 3D State}

\author{Jihai Zhang$^{1}$\thanks{\raggedright Equal contribution. Jihai Zhang's work was done during an internship at Tencent.\\
Code and weights: \href{https://github.com/MajorDavidZhang/PlayGen-2}{\texttt{github.com/MajorDavidZhang/PlayGen-2}}.}\hspace{0.4em},
Yuhan Zhou$^{2}$\footnotemark[1]\hspace{0.4em}, Junyou Li$^{3}$, Deheng Ye$^{3}$, Qi Dou$^{1}$, Yu Cheng$^{4}$ \\[4pt]
\normalfont $^{1}$The Chinese University of Hong Kong \quad $^{2}$Wellington College \\
\normalfont $^{3}$Tencent \quad $^{4}$Nanyang Technological University
}

\hypersetup{
    pdftitle={PlayGen2: Learning Real-Time Playable Worlds with Explicit 3D State},
    pdfauthor={Jihai Zhang, Yuhan Zhou, Junyou Li, Deheng Ye, Qi Dou, Yu Cheng}
}

\iclrfinalcopy
\begin{document}

\maketitle
\pagestyle{plain}

\begin{figure}[!ht]
    \centering
    \includegraphics[width=\linewidth]{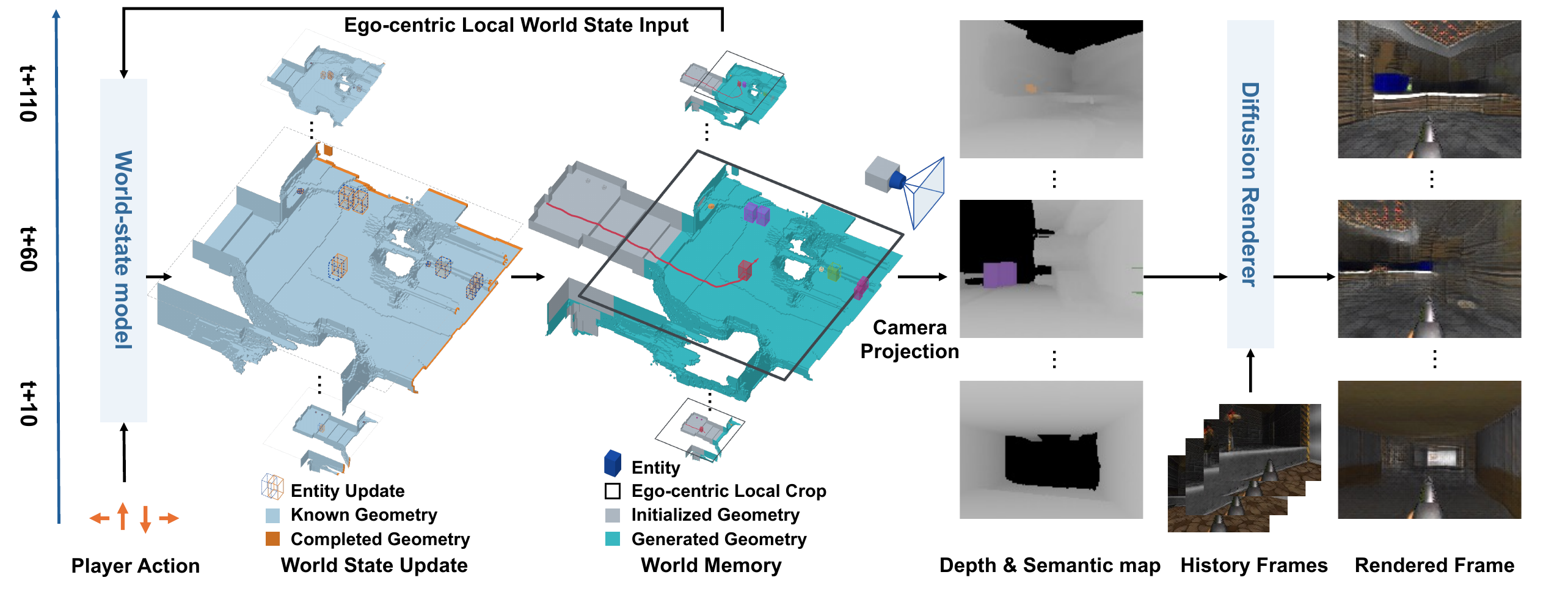}
    \caption{\textbf{PlayGen2 overview.} An action-conditioned world-state
    model updates an ego-centric region of persistent world memory. A separate
    diffusion renderer synthesizes observations from projected depth and
    semantics, actions, and visual history.}
    \label{fig:method-overview}
\end{figure}

\begin{abstract}
    Building a playable world requires more than generating plausible video: it demands responsive control, persistent spatial structure, and coherent interactions with dynamic entities.
    Using generated frames as the sole recurrent world representation ties these requirements to view-dependent appearance synthesis, making it difficult to maintain long-term spatial and temporal consistency in generation and interaction.
    We present PlayGen2, a real-time interactive world model that separates learned 3D state evolution from neural rendering.
    PlayGen2 maintains a persistent world memory comprising static geometry represented by a volumetric occupancy grid and identity-indexed 3D states for the player and other entities.
    Within a fixed-size local region, its world-state model jointly learns action-conditioned entity dynamics, geometry completion, and entity discovery, committing updates to a persistent world memory.
    A separate neural renderer synthesizes first-person observations from depth and semantic projections of the world state, without feeding generated images back into state prediction.
    Using Doom as a testbed, we demonstrate that the model successfully learns world dynamics including geometry-aware player motion and player--NPC interactions. The framework is
      computationally efficient, enabling the complete simulation and
      rendering pipeline to achieve approximately 160\,ms backend latency
      on a MacBook Pro with an M4 Pro chip.
    Beyond single-player interaction, we demonstrate zero-shot deployment on unseen Doom maps, multiplayer interaction through the composition of single-player models, and programmable gameplay through rule-based state interventions. These capabilities demonstrate the transferability and extensibility of the learned game-world model.
    Code and pretrained model weights are publicly available.
\end{abstract}

\section{Introduction}
\label{sec:introduction}

Playing in a generated world requires more than a sequence of plausible images.
A player expects a corridor to remain in place after turning away, a wall to
constrain movement, and nearby entities to evolve while controls remain
responsive. These expectations concern a world that persists and changes under
interaction, not merely the appearance of its next frame. Building a playable
world model therefore requires bringing together low-latency control,
persistent spatial structure, and coherent interactions~\citep{yang2024playgen,garcin2026persist,duan2026liveworld}.

Action-conditioned video generation offers a compelling route toward this
goal~\citep{bruce2024genie}. Systems such as
GameNGen~\citep{valevski2025gamengen} and PlayGen~\citep{yang2024playgen}
demonstrate that neural models can simulate gameplay and support real-time
interaction directly from visual histories. However, when generated
frames serve as the sole recurrent world representation, the model must use
the same view-dependent signal to preserve geometry, track entities, predict
their interactions, and synthesize appearance. Information outside the current
view must be recovered from visual memory~\citep{garcin2026persist, qian2026matrixgame35}, and errors
in generated visual histories can propagate through autoregressive
prediction~\citep{valevski2025gamengen}. Moreover, a pixel-based interface does not directly expose entity states for editing, participant composition, or rule-based interventions~\citep{po2026multigen,huang2026pwm}. 
To address these problems, recent work has begun to separate simulation from rendering~\citep{garcin2026persist,duan2026liveworld,po2026multigen,zhao2026khora}. This separation, however, does not by itself determine how to learn a persistent game world that both evolves under interaction and expands through exploration.

Inspired by the explicit world state maintained by traditional game engines, we organize learned game-world simulation around two complementary processes.
\emph{State transition} predicts how the player and other existing entities evolve, conditioned on player actions and the current world state.
\emph{Incremental world generation} completes previously unknown geometry and introduces newly encountered entities. These processes are coupled through a shared, persistent 3D state: newly generated geometry provides context for subsequent state transitions, while the resulting player motion determines which previously unknown regions need to be generated next. We therefore seek to jointly learn incremental world generation and action-conditioned state transitions, while keeping appearance synthesis separate.

We present \textbf{PlayGen2}, a real-time playable world model that combines a
learned \emph{world-state model} with a \emph{neural renderer}
(Figure~\ref{fig:method-overview}). Inspired by
traditional game engines, we organize the world state into static geometry and dynamic entities.
Static geometry is represented as a volumetric occupancy grid, while dynamic entities have identity-indexed states describing their 3D bounding volumes, semantic classes, and motion attributes. These entities include the player, monsters, projectiles, and items: their states can change even when they are not moving, as when an item is
collected. Together, geometry and entities form a complete world state. We maintain this state in a persistent world memory that retains geometry and identity-indexed entity records beyond the model’s local processing region.

The world-state model is trained on structured game data collected from the game engine, including 3D geometry, entity states, and player actions. To keep neural computation efficient, it operates on a
fixed-size, player-centered 3D volume read from world memory. Within this
volume, it advances existing entities, completes previously unknown geometry,
and discovers newly encountered entities. State transitions are conditioned on
user actions, entity histories, and geometric context. The resulting updates are written back to world
memory, so generated regions remain available on revisits and entity records
remain stored outside the active volume.

The neural renderer converts the predicted world state into RGB video.
We project the occupancy grid and entity geometry into depth and semantic maps
from the player's camera viewpoint, providing explicit spatial conditions for
frame generation. The separately trained renderer combines these conditions with
recent RGB history to synthesize appearance and maintain visual continuity.
Generated frames update only this RGB history and never feed back
into world-state prediction.

We use Doom as a controlled testbed. On a MacBook Pro with an M4 Pro chip,
our Doom prototype achieves approximately \textbf{160\,ms backend latency},
supporting interactive exploration with dynamic entities. The model learns general world dynamics including collision-aware player navigation and player-NPC interaction. We further evaluate transfer to unseen Doom maps and training-free extensions to multiplayer and programmable gameplay.
Our contributions are threefold:
\begin{itemize}[leftmargin=*]
\item \textbf{A unified framework for learned 3D worlds.}
We introduce a framework for learning world evolution in an explicit
3D representation space comprising volumetric geometry and identity-indexed
entity states. The world-state model jointly learns geometry completion,
entity discovery, and action-conditioned dynamics within a fixed-size
local region, committing updates to persistent world memory while
remaining decoupled from neural rendering.
\item \textbf{Learned game dynamics for real-time play.}
Using Doom as a testbed, we demonstrate that the model successfully learns world dynamics including geometry-aware player motion and player--NPC interactions. Our proposed framework is
computationally efficient, enabling the complete simulation and
rendering pipeline to run in real time on a consumer laptop
with low end-to-end latency.
\item \textbf{Generalization and composition for extensible gameplay.}
We illustrate the generalizability and extensibility of the framework through three applications: zero-shot continuation on unseen Doom maps, multiplayer interaction through the composition of single-player models, and programmable gameplay through state-level interventions. These examples demonstrate how the learned game-world dynamics can be reused and composed beyond the original training setting.
\end{itemize}

\section{Related Work}
\label{sec:related-work}

\subsection{Video-Based Interactive World Models}

Video-based game world models have progressed from demonstrating learned
gameplay to improving controllability, efficiency, and long-horizon consistency.
GameNGen and PlayGen show that action-conditioned video prediction can support
real-time gameplay and interactive
mechanics~\citep{valevski2025gamengen,yang2024playgen}.
Matrix-Game extends image-conditioned generation with keyboard and mouse
control~\citep{zhang2025matrixgame}, while few-step streaming generation and
causal distillation improve interactive efficiency in Matrix-Game 2.0 and
LingBot-World~\citep{he2025matrixgame2,robbyant2026lingbotworld1}.
Beyond speed, recent work addresses what the model remembers and how users
interact: Matrix-Game 3.0 retrieves camera-relevant
history~\citep{wang2026matrixgame3}, Matrix-Game 3.5 adds geometry-aligned
patch memory and dynamic-subject references~\citep{qian2026matrixgame35},
and LingBot-World 2.0 introduces agent-mediated interactions beyond
navigation~\citep{gao2026lingbotworld2}.
However, improved local visual quality does not necessarily imply persistent
world consistency. WorldProbe finds that recent video world models remain
vulnerable to long-range revisit inconsistencies, entity drift, and incoherent
scene evolution under closed-loop spatial
exploration~\citep{wang2026worldprobe}.
These findings motivate maintaining world state independently of generated
observations, rather than relying on visual history alone to preserve the
evolving environment.

\subsection{Explicit-State World Models and Generative Rendering}

Explicit-state approaches separate world evolution from observation synthesis,
but differ in how the world is advanced. PERSIST learns voxel-world evolution
and camera dynamics, while LiveWorld advances out-of-sight entities through
local videos lifted into dynamic point
clouds~\citep{garcin2026persist,duan2026liveworld}.
Shared state also supports editable multiplayer worlds in MultiGen,
population-scalable simulation in Khora, and learned joint-action transitions
in multiplayer Snake in MASS~\citep{po2026multigen,zhao2026khora,cai2026mass}.
A rapidly growing branch instead uses \emph{code-defined worlds rendered by
video models}: executable programs maintain state and determine interaction
outcomes, while video models synthesize their appearance.
Code World Model, CoDeR, and Programmable World Model construct executable
worlds through agents and programs~\citep{chen2026codeworldmodel,fang2026coder,huang2026pwm}.
Magpie, Matrix-Game 4.0, and StreamRender-H3 couple game execution with neural
rendering for interactive play~\citep{zhan2026magpie,zhang2026matrixgame4,lu2026streamrenderh3},
while AgentGarten uses simulator-backed neural-rendered environments for agent
learning~\citep{chi2026agentgarten}.
In this code-driven branch, executable state provides a persistent interface
for rendering, while programs determine the underlying world evolution.

PlayGen2 explores a complementary route: the structured world state itself is
generated and advanced by a learned world-state model, rather than by executing
code-defined dynamics. The model jointly learns geometry completion, entity
discovery, and action-conditioned state transitions from structured game data;
a separate renderer synthesizes observations from depth and semantic projections
of the predicted state. Fixed-size local updates and persistent world memory
support efficient exploration, while the explicit state remains accessible for
composition and intervention. Optional rules extend this learned simulation
rather than supplying its base dynamics. Thus, PlayGen2 learns game-world
evolution as well as its visual realization, while retaining the explicit
interface used by code-driven systems.

\section{Method}
\label{sec:method}
PlayGen2 maintains a learned simulation loop over explicit 3D state
  and a separate rendering loop conditioned on projected world state
  and RGB history (Figure~\ref{fig:method-overview}).
  The two models are trained separately using structured game-state
  and action sequences for simulation, and aligned state projections
  and RGB observations for rendering.

  \subsection{World State and System Overview}
  \label{sec:world-state}

  The simulation state comprises a volumetric occupancy map $M_t$,
  an availability mask $K_t$, player state $e_t$, and an
  identity-indexed object table $\mathcal{O}_t$:
  \begin{equation}
      \mathcal{S}_t=(M_t,K_t,e_t,\mathcal{O}_t),
      \qquad
      o_t^i=(c_t^i,\mathbf{p}_t^i,h_t^i,r_t^i,\mathbf{v}_t^i).
      \label{eq:world-state}
  \end{equation}
  Here, $i$ indexes an object in $\mathcal{O}_t$,
  $c_t^i$ is its semantic class,
  $\mathbf{p}_t^i \in \mathbb{R}^3$ is its base position,
  $h_t^i$ and $r_t^i$ are its height and horizontal radius,
  and $\mathbf{v}_t^i \in \mathbb{R}^3$ is its velocity.
  The player state $e_t$ contains the corresponding continuous
  attributes together with its orientation.
  The player action $a_t$ controls the transition from $t$ to $t+1$.
  The availability mask $K_t$ indicates whether a voxel has an
  assigned occupancy value, supplied at initialization or committed
  from an earlier prediction; unavailable voxels are not treated
  as empty space.

  Persistent world memory retains committed geometry and entity
  records, while the network processes a player-centered 3D crop
  and a sliding window of identity-aligned history.
  Objects outside this volume are stored but not continuously simulated.
  Before encoding, all historical positions are expressed relative
  to the same current crop origin.
The world-state model uses the current state and its history
  to predict state updates.
  For rendering, the updated state is projected into depth
  and semantic maps that condition RGB generation.
  RGB history is used only by the renderer; generated images
  never enter the world-state prediction loop.

\subsection{Geometry-Aware World Evolution}
  \label{sec:layout}
\begin{figure}[t]
      \centering
      \includegraphics[width=\linewidth]{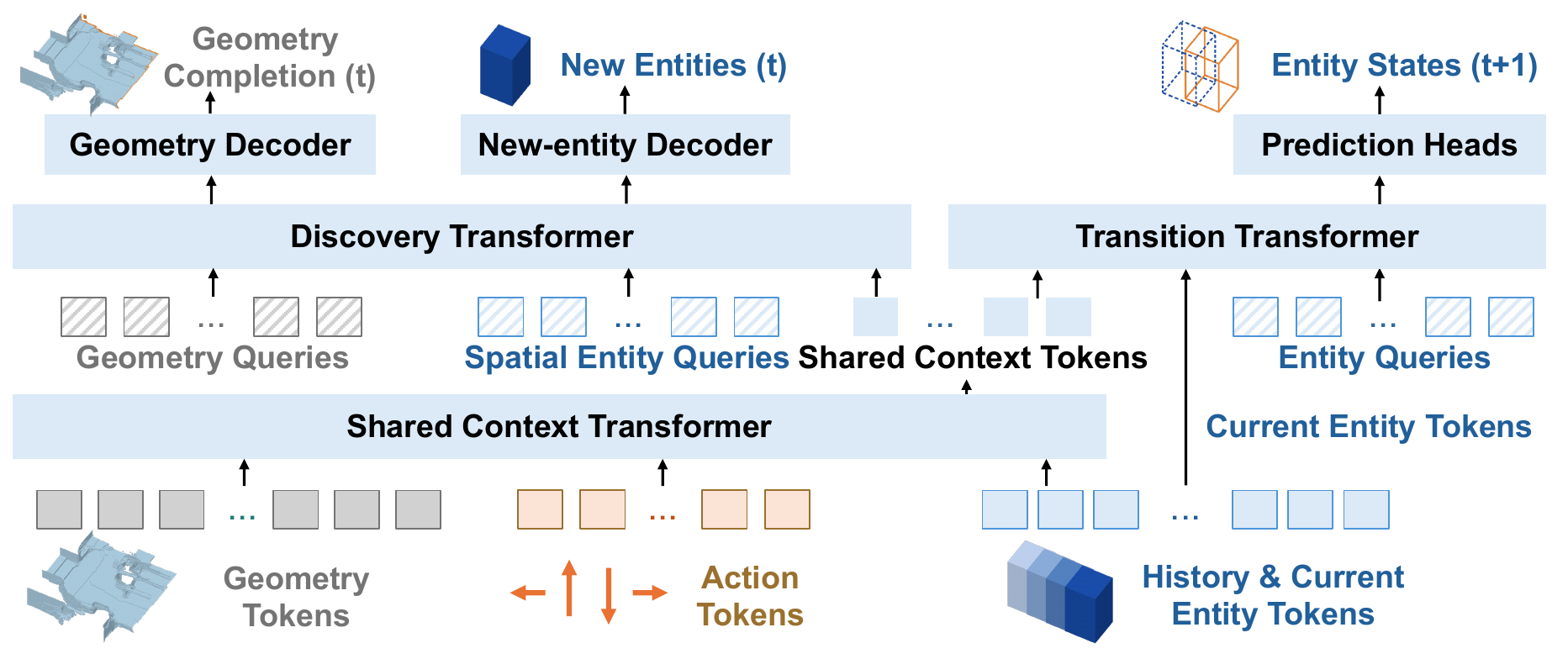}
      \caption{\textbf{World-state model architecture.}
      The Shared Context Transformer encodes geometry, entity histories, and actions.
      The Discovery Transformer processes Geometry Queries and Spatial Entity Queries
      to complete geometry and generate new entities at time $t$.
      The Transition Transformer predicts player and entity states at $t+1$,
      additionally attending to Current Entity Tokens that include newly accepted entities.
      The state-commit and re-embedding path is omitted for clarity.}
      \label{fig:world-state-architecture}
  \end{figure}

  Figure~\ref{fig:world-state-architecture} outlines the world-state model.
  A 3D convolutional encoder processes the local occupancy map, availability
  mask, and a per-voxel vertical coordinate relative to the player. The Shared Context Transformer combines geometry tokens with player and object histories, and action
  information to produce shared context $\mathcal{C}_t$, which supports incremental world generation at time $t$
  and state transitions from $t$ to $t+1$.
  
  \textbf{Incremental world generation.}
  The Discovery Transformer processes two types of queries conditioned on
  $\mathcal{C}_t$. Geometry Queries are associated with spatial locations in the
  coarse feature volume. Their outputs pass through the Geometry Decoder,
  which reconstructs voxel occupancy using 3D upsampling and encoder skip connections.
  Spatial Entity Queries propose new entities around candidate 3D reference
  positions, or anchors, derived from the available world state.
  Each query combines a learned embedding with encodings of its anchor position
  and proposal metadata. The New-entity Decoder uses the resulting query features
  and geometry encoder features to predict objectness, semantic class,
  anchor-relative position, size, and velocity.
  Anchor construction, matching, and proposal filtering are detailed in
  Appendix~\ref{app:layout}.

  \textbf{Geometry-conditioned state transitions.}
  Entity Queries predict the next states of the player and active objects.
  For each non-player object, query conditioning combines its current
  state, position differences between consecutive available history
  frames, and local geometry sampled from the encoder.
  Attention over the sampled geometry features provides
  object-specific spatial context. Player motion accesses geometry, actions,
  and history through the shared context.
 The Transition Transformer updates Entity Queries through
  self-attention and cross-attention to the shared context
  $\mathcal{C}_t$ concatenated with Current Entity Tokens.
  These current-state tokens include newly accepted entities,
  making them available to state prediction without recomputing
  the shared context.
  The updated queries are passed to prediction heads to produce
  the next player and object states. Positions are predicted as residual updates:
  \begin{equation}
      \hat{\mathbf{p}}_{t+1}^i
      =\mathbf{p}_t^i+f_{\mathrm{xyz}}(\mathbf{h}_{t,\mathrm{trans}}^i),
      \label{eq:world-displacement}
  \end{equation}
  where $\mathbf{h}_{t,\mathrm{trans}}^i$ is the output feature
  of the Transition Transformer for entity $i$, and
  $f_{\mathrm{xyz}}$ maps this feature to a predicted 3D displacement. The player uses the same residual-position form. Separate heads predict
  class, size, velocity, and player orientation. No runtime collision correction is applied.

  \textbf{Update order and persistent memory.}
  Incremental world generation completes the local region at the
  current, already committed player pose before state transition
  predicts the next player and object states.
  Let $D_\theta$ and $T_\theta$ denote the generation and transition
  branches, including their respective output decoders and heads.
  Both reuse the shared context $\mathcal{C}_t$ and geometry encoder
  features $\mathcal{F}_t$.
  Here, $\mathcal{A}_t$ denotes the anchors and metadata used to
  construct Spatial Entity Queries.
  $\mathcal{H}_t^+$ denotes entity histories updated with accepted
  new entities. These histories provide the current-state and
  motion features used to encode Current Entity Tokens and
  condition Entity Queries.
  Using $-$ and $+$ for states before and after incremental world generation:
  \begin{equation}
      \mathcal{S}_t^+
      = \operatorname{Commit}\bigl(
          \mathcal{S}_t^-,D_\theta(\mathcal{C}_t,\mathcal{F}_t,\mathcal{A}_t)
        \bigr),
      \qquad
      (\hat e_{t+1},\hat{\mathcal{O}}_{t+1})
      = T_\theta(\mathcal{C}_t,\mathcal{F}_t,\mathcal{H}_t^+).
      \label{eq:world-update}
  \end{equation}
  Generated occupancy is written only to unavailable voxels,
  using the same crop origin as the input geometry.
  This ordering decouples current-region generation and memory
  commitment from same-step player-position prediction, avoiding
  misalignment between generated voxels and their writeback locations.
  The predicted player pose determines the local region processed
  at the next step instead.
  Accepted entities enter history and become eligible for state
  transition. The original context and geometry features are reused
  without re-encoding the completed map.
  The next state retains the committed geometry and incorporates
  the predicted player and object updates.
  Newly introduced entity states are teacher-forced during training
  and supplied by accepted predictions during rollout.
  
\textbf{Training and robustness.}
  We jointly supervise incremental world generation at time $t$
  and entity state transitions from $t$ to $t+1$.
  Newly introduced entities are excluded from the shared context,
  while their ground-truth current states are supplied to the
  transition branch during training.
  Unknown geometry receives focal and volumetric Dice losses~\citep{milletari2016vnet},
  with a separate reconstruction loss on known regions.
  New-entity losses separate sparse objectness prediction from
  semantic classification. For non-item objects, phase-balanced
  position losses balance the contributions of moving, stopping,
  and stationary examples.

  Random history truncation trains the model to handle incomplete
  histories, including entities with only one available frame.
  Temporally correlated position perturbations are applied to
  stationary item histories, with clean targets, to train recovery
  from accumulated drift.
  Architecture, assignment, objectives, and rollout details appear
  in Appendix~\ref{app:layout}.
  
 \subsection{State-Conditioned Neural Rendering}
  \label{sec:renderer}

  Let $x_t$ denote the RGB observation at time $t$.
  The renderer generates the next observation conditioned on
  the projected world state, an $L$-frame RGB history, and
  the player action:
  \begin{equation}
      x_{t+1}\sim p_\phi\bigl(x\mid
          c_{t+1},x_{t-L+1:t},a_t\bigr),
      \qquad c_{t+1}=\Pi(\mathcal{S}_{t+1}^-).
      \label{eq:render-conditional}
  \end{equation}
  Here, $\Pi$ uses occupancy ray casting and entity bounding-box projection
  to produce semantic and normalized depth maps from the player's viewpoint,
  together with validity masks (Appendix~\ref{sec:renderer-projection}).
  These scene conditions specify the visible structure, while
  RGB history provides appearance information and supports
  temporal continuity.

  We use a latent diffusion Transformer with a frozen,
  domain-trained variational autoencoder.
  Separate encoders transform the latent representations of
  RGB history and the scene conditions into RGB-history tokens
  and scene-condition tokens, respectively.
  Each denoising block cross-attends to both sets of tokens.
  Diffusion noise is applied only to the target latent.

  After training with ground-truth RGB history, we fine-tune
  the renderer by mixing ground-truth and cached autoregressively
  generated RGB histories and applying RGB-history dropout.
  Generated histories are collected offline from a frozen
  renderer checkpoint; the target remains the recorded next frame.
  RGB-history dropout masks out the entire history for a subset
  of training examples while retaining scene conditions and actions.
  This training exposes the renderer to imperfect or absent
  RGB history while retaining supervision for temporal continuity
  from ground-truth RGB history.
  Architecture, training objectives, history construction,
  and sampling details are provided in Appendix~\ref{app:renderer}.

\section{Experiments}
\label{sec:experiments}

We evaluate learned game dynamics through player–geometry and player–NPC interactions, assess long-horizon rendering and world generation on seen and unseen maps, and measure end-to-end inference efficiency. We also demonstrate multiplayer composition and programmable gameplay without retraining. Together, these experiments show that PlayGen2 supports real-time interactive world simulation with transferable dynamics and an extensible state interface.

\label{sec:exp-setup}

\textbf{Training data and compute.}
We train on approximately 500 hours of gameplay automatically collected
exclusively on Freedoom2 MAP01 using a rule-based controller in
ViZDoom~\citep{kempka2016vizdoom}. The dataset totals roughly 500\,GB. The world-state
model and rendering model contain 41.2\,M and 29.1\,M parameters, respectively;
the latter excludes the fixed 55.3\,M-parameter VAE. RGB frame resolution is $160\times120$. All models are trained from scratch. The total training budget
is approximately three days on four NVIDIA A100 GPUs (288 GPU-hours).

\textbf{Evaluation setting.}
Seen-map evaluation uses Freedoom2 MAP01. Unseen-map evaluation uses seven levels from Scythe, a community-created Doom II map pack. Starting positions are fixed before evaluation, and the same starting configurations and controller settings are used across global and local initialization. For learned-world rollout, \emph{global}
initialization supplies the full static map, whereas \emph{local}
initialization supplies only the initial neighborhood and requires learned
completion during exploration. Both initialization modes provide local entity histories only at the start of rollout; subsequent entity states and newly discovered entities are predicted by the model. We evaluate each map from multiple predefined starting positions using a rule-based stochastic exploration controller to select actions. A failure occurs when more than half of the player's bounding box overlaps occupied geometry. The player respawns after a
  failure. We evaluate 64 exploration trajectories for each combination of map
  familiarity (seen/unseen) and map initialization (global/local).
  Each trajectory lasts 10 simulated minutes, yielding 256 trajectories
  in total.
 Visual-quality evaluation uses BF16 inference and eight renderer denoising
  steps by default. Each simulation step corresponds to three engine tics
  at 35\,Hz, yielding 700 steps per simulated minute. The 1-, 5-, and
  10-minute stages below refer to simulated time rather than wall-clock
  duration. Checkpoints and detailed protocols are provided in
  Appendix~\ref{app:experiments}.

\subsection{Learned Game Dynamics and Control}
\label{sec:exp-world-state}

We measure player--geometry interaction using \emph{contact error rate}
(CER), the fraction of resolved obstacle encounters and landings that end
in severe penetration. CER aggregates resolved encounters across respawns up to
  each evaluation horizon.
RGB action accuracy (ActAcc)~\citep{yang2024playgen} measures agreement between commanded actions
and actions recognized from generated RGB, assessing control through both
dynamics and rendering.

\begin{table}[!htb]
    \centering
    \caption{\textbf{Interaction over time.} CER is cumulative
    through each time horizon; ActAcc measures action recognition from
    generated RGB. Global/local denotes map
    initialization.}
    \label{tab:exp-world-state}
    \small
    \setlength{\tabcolsep}{4pt}
    \begin{tabular*}{\linewidth}{@{\extracolsep{\fill}}lrrrrrr@{}}
        \toprule
        & \multicolumn{3}{c}{CER (\%) $\downarrow$}
        & \multicolumn{3}{c}{ActAcc (\%) $\uparrow$} \\
        \cmidrule(lr){2-4}\cmidrule(l){5-7}
        Setting & 1 min & 5 min & 10 min & 1 min & 5 min & 10 min \\
        \midrule
        Seen/global   & 2.0 & 4.8 & 5.3 & 93.7 & 85.4 & 83.9 \\
        Seen/local    & 3.1 & 4.3 & 6.7 & 91.2 & 87.8 & 82.1 \\
        Unseen/global & 3.7 & 4.4 & 5.2 & 94.6 & 84.1 & 85.0 \\
        Unseen/local  & 3.6 & 5.6 & 6.8 & 93.3 & 88.2 & 85.6 \\
        \bottomrule
    \end{tabular*}
\end{table}

Table~\ref{tab:exp-world-state} demonstrates sustained action-conditioned
  control over ten-minute rollouts, achieving ActAcc of 82.1--85.6\%
  and cumulative CER of 5.2--6.8\%.
  Comparable performance on seen and unseen maps shows that the model
  learns transferable player dynamics rather than memorizing interactions
  within the training map. Both CER and ActAcc remain
  relatively stable between five and ten minutes, indicating sustained
  control and physical interaction over long-horizon rollouts. Together, these results demonstrate the framework's ability to maintain controllable, geometry-aware interaction
  over long horizons and across new environments.

\textbf{Monster--player interaction.}

\begin{figure}[!htb]
    \centering
    \includegraphics[width=\linewidth]{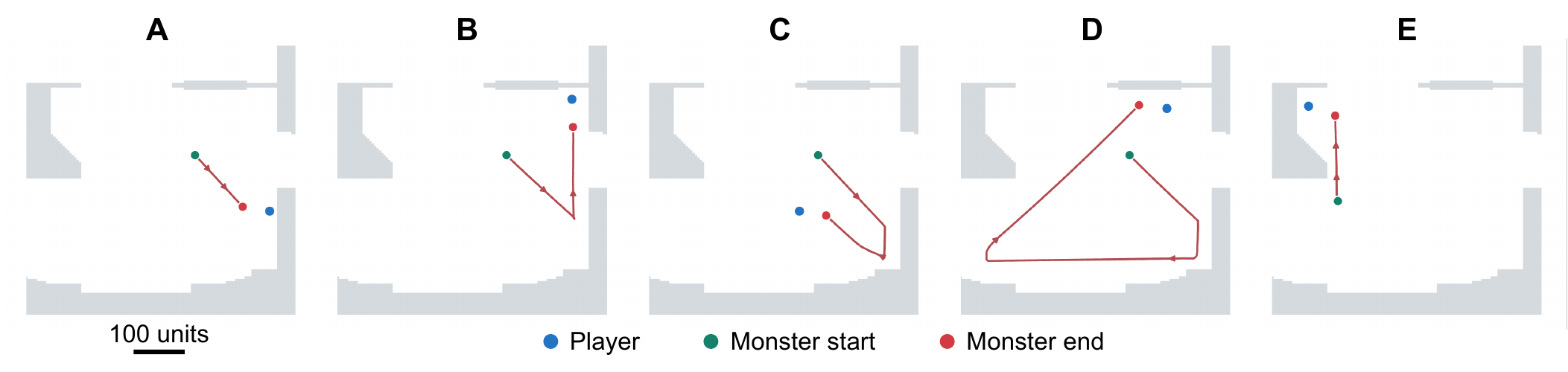}
    \caption{\textbf{Learned monster motion toward a stationary player.}
    Blue marks the player; green and red mark the monster's start and end.
    Arrows indicate temporal direction. These selected successful proximity
    trajectories use a 300-step budget, distinct from the 200-step success
    statistic in the text: D reaches the threshold at step 211, and E is a
    separately initialized wall-side example. The model updates monster
    states without scripted pursuit or runtime collision correction.}
    \label{fig:exp-monster-motion}
\end{figure}

  In Doom, monsters pursue the player as part of their combat behavior.
  We evaluate randomly sampled player locations around an observed
  Zombieman initialization, keeping the player stationary during each
  rollout. Pursuit is considered successful when the monster reaches
  a horizontal body-surface distance less than 30 world units from
  the player. The model achieves a 55\% pursuit success rate within
  200 steps. Figure~\ref{fig:exp-monster-motion} illustrates successful
  model-generated approaches toward different player locations,
  demonstrating that our model learns game world dynamics
  that encompass player--NCP interactions and reasonable NPC behavior.

Importantly, these interactions extend beyond the camera's field
  of view. PlayGen2 updates NPC states throughout its active 3D
  simulation volume, regardless of whether they are currently visible.
  A monster can therefore continue pursuing the player and approach
  from behind even while the player is looking elsewhere.
  By decoupling entity dynamics from visual observation, our framework
  allows off-screen entities to remain active participants in gameplay:
  turning away changes what the player sees, not whether the surrounding
  world continues to evolve. This capability supports a more coherent
  and engaging interactive experience.

\subsection{Long-Horizon Rendering and World Generation}
\label{sec:exp-comparison}
We evaluate long-horizon visual quality under global and local initialization on seen and
  unseen maps. For each trajectory, we sample three 32-frame clips from a one-minute window
  near each evaluated horizon. All horizons refer to simulated time. Under global
  initialization, the explicit player state allows us to extract camera trajectories from
  rollouts driven by a rule-based stochastic exploration policy and replay them in the
  ViZDoom engine to obtain pose-aligned reference videos. Under local initialization, the
  model generates previously unknown geometry, so no unique aligned engine reference exists.
  We therefore report only FID and FVD for this setting, using a fixed Freedoom2 reference
  pool.

  We compare with PlayGen~\citep{yang2024playgen} on Freedoom2 MAP01, the training map shared
  by both systems. We convert the exploration trajectories into action sequences and use them
  to drive PlayGen. Notice that PlayGen do not have the capabilities to generalize on unseen maps.  We report PlayGen results at one and five minutes; by five minutes, its
  generated videos exhibit substantial visual degradation.

\begin{figure}[!htb]
    \centering
    \includegraphics[width=\linewidth]{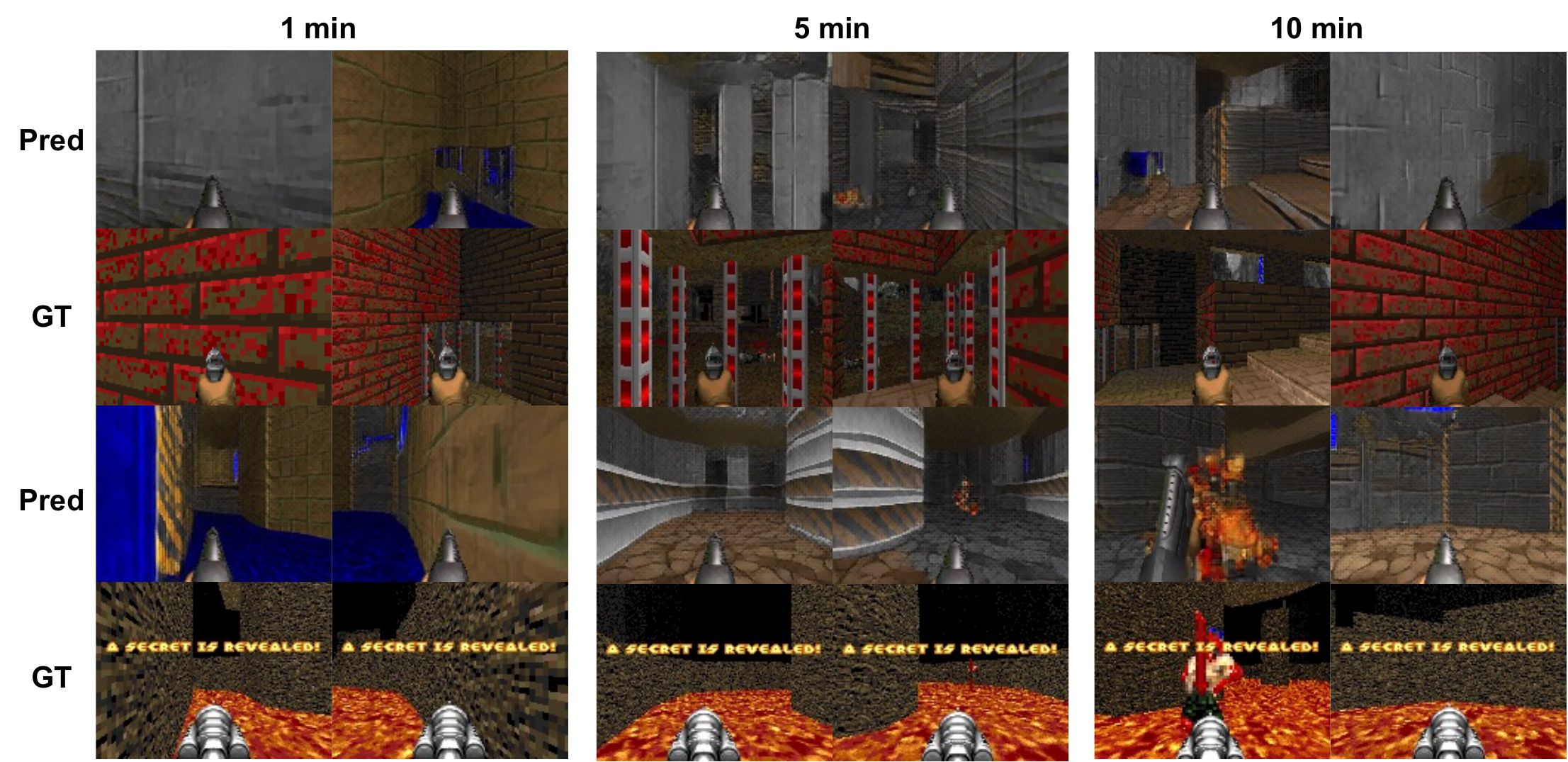}
    \caption{\textbf{Long-run rendering on unseen maps.}Two trajectories sampled at 1, 5, and 10 simulated minutes.
  Each cell shows the first and last frames of a 32-frame clip,
  with generated images above pose-aligned engine references.}
    \label{fig:exp-quality-long}
\end{figure}

\begin{table}[!htb]
    \centering
    \caption{\textbf{Visual quality over time.} Global and Local denote PlayGen2's world-memory initialization.
  PlayGen receives trajectory-derived actions and is compared only on seen maps.
  Local initialized generation is evaluated against unpaired references.
  Times denote simulated minutes; N/A indicates an inapplicable setting.}
    \label{tab:exp-main}
    \label{tab:exp-renderer-long}
    \label{tab:exp-local-quality}
    \small
    \setlength{\tabcolsep}{2pt}
    \begin{tabular*}{\linewidth}{@{\extracolsep{\fill}}llcrrrrrrr@{}}
        \toprule
        & & & \multicolumn{5}{c}{Global}
        & \multicolumn{2}{c}{Local} \\
        \cmidrule(lr){4-8}\cmidrule(l){9-10}
        Setting & Method & Min & PSNR $\uparrow$
        & SSIM $\uparrow$ & LPIPS $\downarrow$
        & FID $\downarrow$ & FVD $\downarrow$ & FID $\downarrow$ & FVD $\downarrow$ \\
        \midrule
        \multirow{5}{*}{Seen} & \multirow{3}{*}{PlayGen2}
        & 1 & 18.3 & 0.40 & 0.41 & 223.1 & 1302.7 & 233.4 & 1720.6 \\
        & & 5 & 19.4 & 0.39 & 0.42 & 237.4 & 1298.3 & 253.2 & 1638.5 \\
        & & 10 & 17.3 & 0.39 & 0.44 & 250.3 & 1261.9 & 243.6 & 1673.1 \\
        \cmidrule(l){2-10}
        & \multirow{2}{*}{PlayGen}
        & 1 & 19.3 & 0.39 & 0.38 & 210.9 & 1097.5
        & \multicolumn{2}{c}{\multirow{2}{*}{N/A}} \\
        & & 5 & 16.1 & 0.29 & 0.69 & 373.4 & 2386.9
        & \multicolumn{2}{c}{} \\
        \midrule
        \multirow{3}{*}{Unseen} & \multirow{3}{*}{PlayGen2}
        & 1 & 16.6 & 0.28 & 0.65 & 249.4 & 1747.8 & 247.7 & 1744.4 \\
        & & 5 & 16.5 & 0.29 & 0.63 & 240.5 & 1605.3 & 240.8 & 1736.3 \\
        & & 10 & 16.1 & 0.26 & 0.66 & 246.1 & 1713.5 & 239.0 & 1688.0 \\
        \bottomrule
    \end{tabular*}
\end{table}

\textbf{Seen-map comparison.}
 The two systems achieve comparable visual quality at one minute. By five minutes, however,
  PlayGen's quality deteriorates substantially, whereas PlayGen2 retains substantially more
  stable scores and performs better on all five metrics. Across the one-, five-, and ten-
  minute stages, PlayGen2's LPIPS remains within 0.41--0.44. These results support the
  framework's ability to sustain visual quality during long-horizon generation.

\textbf{Unseen maps.}
  Figure~\ref{fig:exp-quality-long} shows zero-shot rendering on unseen Scythe maps. The
  examples illustrate that PlayGen2 follows the supplied spatial structure while synthesizing
  appearance in its learned Freedoom2 style. Because depth and semantic conditions do not
  specify the native Scythe textures, reference-based metrics reflect both structural
  discrepancies and appearance differences; we therefore interpret them alongside qualitative
  results and their variation over time. PSNR remains within 16.1--16.6\,dB and LPIPS within
  0.63--0.66 across the sampled stages, showing limited variation over ten simulated minutes
  and supporting transfer to layouts absent from training.

\begin{figure}[!htb]
    \centering
    \includegraphics[width=\linewidth]{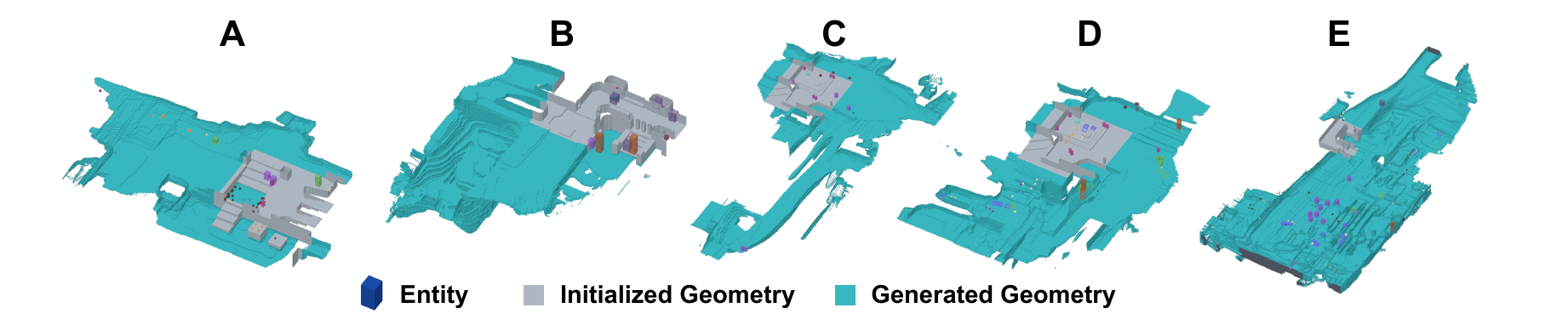}
    \caption{\textbf{World maps generated from local initialization.}
    Five rollout snapshots at 1 minute (A--C) and 10 minutes (D--E).
  A starts from a local neighborhood of the training map;
  B--E start from unseen maps.
  Cutaway views distinguish initialized geometry (gray),
  generated extensions (cyan), and class-colored entity boxes.
  The player is omitted.}
    \label{fig:exp-map-generation}
\end{figure}

\textbf{Generating local worlds.}
Starting from a local neighborhood, PlayGen2 incrementally generates
  geometry and introduces entities as the player explores.
  FID and FVD remain broadly stable across the evaluated horizons,
  supporting sustained visual quality as the generated world expands.
  Figure~\ref{fig:exp-map-generation} shows the resulting spatial
  layouts, distinguishing initialized geometry from generated
  extensions and newly introduced entities.
  Despite training on a single map, the model generates plausible
  geometry and entity configurations from both seen- and unseen-map
  initializations. These results indicate that PlayGen2 learns
  transferable spatial priors for game-world structure:
  the learned regularities support world expansion beyond the
  specific training layout and can be applied to new spatial contexts
  without retraining.

\subsection{Extensible Gameplay}
\label{sec:exp-extensions}
PlayGen2's explicit state interface enables participant synchronization
  and programmable rules on top of learned dynamics. We demonstrate
  multiplayer composition and programmable interaction using models
  trained exclusively on single-player gameplay, without updating their
  weights.

\textbf{Multiplayer composition.}
  We compose single-player model instances by exchanging player states
  and representing the other participant as a Zombieman.
  Each player supplies independent actions and maintains a separate
  RGB history. Both players' motion is predicted by learned
  action-conditioned transitions without runtime collision correction.
  Following the opponent-presence evaluation of
  MultiGen~\citep{po2026multigen} as closely as possible,
  we assess whether generated observations depict the opponent
  consistently with reference visibility.
  PlayGen2 achieves 96.3\% accuracy, 90.5\% precision, and 95.0\% recall,
  with opponent visibility as the positive class
  (Table~\ref{tab:exp-multiplayer}).
  Synchronized views and trajectories for approach, departure, and
  following are shown in Figure~\ref{fig:exp-multiplayer-qualitative};
  setup details are provided in Appendix~\ref{app:exp-extensions}.
  These results demonstrate that models trained for single-player
  interaction can be composed into a multiplayer experience through
  their shared state interface, without multiplayer-specific training.

\begin{figure}[!htb]
    \centering
    \includegraphics[width=\linewidth]{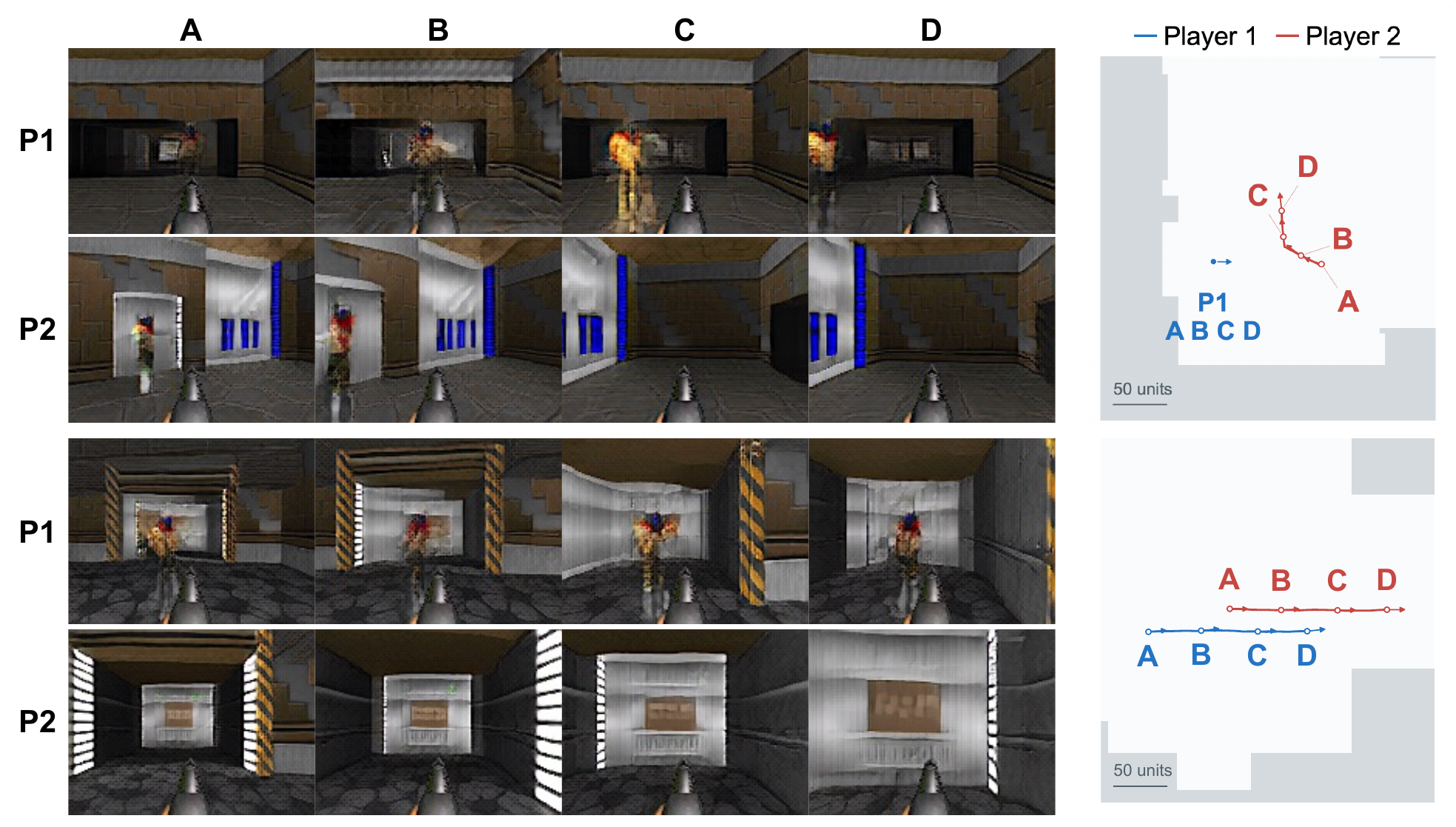}
    \caption{\textbf{Paired multiplayer rollouts in Freedoom2.} Synchronized
    views and trajectories (P1: blue; P2: red); A--D mark displayed times.
    Top: P1 receives idle actions as P2 approaches and leaves its view.
    Bottom: P1 follows P2. Both players use learned transitions and separate
    rendering histories, with no runtime collision correction. Setup details
    are in Appendix~\ref{app:exp-extensions}.}
    \label{fig:exp-multiplayer-qualitative}
\end{figure}

\textbf{Programmable interaction.}
  Figure~\ref{fig:exp-extensions} demonstrates rule-based health loss
  and death implemented through explicit state interventions,
  with feedback overlaid on generated observations.
  The state interface allows these rules to be specified and modified
  directly, rather than requiring the neural models to learn each
  rule from additional examples.
  The learned dynamics and renderer are reused, while selected
  gameplay events are governed by explicit code.
  This combination supports precise control over chosen mechanics
  without replacing the learned simulation, allowing the same
  trained system to accommodate new interaction rules.

\begin{table}[!htb]
    \centering
    \begin{minipage}[t]{0.4\linewidth}
        \vspace{0pt}
        \centering
        \caption{\textbf{Multiplayer opponent-presence consistency (\%).} Opponent
        visible is the positive class.}
        \label{tab:exp-multiplayer}
        \small
        \setlength{\tabcolsep}{3pt}
        \begin{tabular*}{\linewidth}{@{\extracolsep{\fill}}lrrr@{}}
            \toprule
            Method & Acc. $\uparrow$ & Prec. $\uparrow$ & Recall $\uparrow$ \\
            \midrule
            MultiGen & 75.4 & 88.1 & 65.1 \\
            PlayGen2 & 96.3 & 90.5 & 95.0 \\
            \bottomrule
        \end{tabular*}
    \end{minipage}\hfill
    \begin{minipage}[t]{0.55\linewidth}
      \vspace{0pt}
      \centering
      \makeatletter
      \def\@captype{figure}
      \makeatother
      \includegraphics[width=\linewidth]{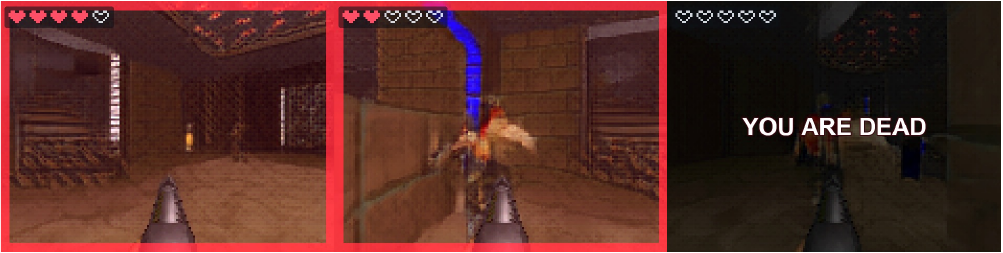}\par
      \caption{\textbf{Programmable health and death.}
      Rule-based damage reduces player health and triggers death.}
      \label{fig:exp-extensions}
  \end{minipage}
\end{table}

\subsection{Interactive Efficiency}
\label{sec:exp-efficiency}
The complete simulation--rendering pipeline is able to run locally on a MacBook Pro
(M4 Pro). With 8 denoising steps, serial execution achieves 4.98 FPS at
200.6\,ms mean per-frame backend latency.Separate profiling measures approximately
75/90/30\,ms for world-state updates, latent rendering, and VAE decoding,
respectively. Overlapping world updates and
rendering on the same device raises throughput to 7.16 FPS (+44\%),
at 275.7\,ms latency. These results demonstrate that our world-state modeling makes a complete learned 3D simulation and rendering loop computationally tractable at interactive rates on consumer hardware.

\FloatBarrier

\section{Conclusion}

We presented PlayGen2, a playable world model that decouples learned
3D world-state evolution from neural rendering. By explicitly representing
static geometry and dynamic entities, the framework integrates geometry
completion, entity discovery, and action-conditioned dynamics within
persistent world memory. Our Doom experiments demonstrate learned
player--geometry and player--NPC interactions, sustained visual quality
over long-horizon rollouts, and interactive execution on a consumer laptop.
Beyond the training environment, the same framework supports zero-shot
deployment on unseen maps, multiplayer composition, and programmable
gameplay without retraining. Together, these results establish explicit
3D world state as a practical foundation for unifying learned simulation,
neural rendering, and flexible interaction in efficient, extensible,
and playable game worlds.

\bibliography{iclr2027_conference}
\bibliographystyle{iclr2027_conference}

\clearpage
\appendix
\section{World-State Model Details}
\label{app:layout}

This appendix specifies the representation, architecture, supervision, and
memory operations underlying Section~\ref{sec:layout}.

\subsection{Persistent World State}
\label{sec:layout-state}

We use the world state defined in Equation~\ref{eq:world-state}. Entity positions
refer to their bases, and player orientation is encoded as
$(\sin\theta_t,\cos\theta_t)$. The object table includes moving entities and
stationary interactable items. Identity-aligned slots and validity masks
distinguish missing observations from observed stationary states.

The network operates on an axis-aligned crop centered approximately on the
current player position, without rotating the map into the viewing direction.
The crop covers $800\times800\times200$ world units with a voxel width of five
units, giving a $40\times160\times160$ volume in $Z,Y,X$ order. Let $V_t^-$ and
$K_t^-$ denote its occupancy and availability before scene completion. A zero
in $K_t^-$ denotes unknown geometry, not empty space. At inference, available
voxels can come from initialization or earlier predictions.

We retain four history frames and at most 24 active object slots. Historical
positions share the current crop origin. For continuous attributes
$\mathbf{x}=(x,y,z,h,r,v_x,v_y,v_z)$, network inputs use the elementwise transform
\begin{equation}
    \bar{\mathbf{x}}=2\frac{\mathbf{x}-\mathbf{l}}{\mathbf{u}-\mathbf{l}}-1,
    \label{eq:layout-normalization}
\end{equation}
where $\mathbf{l}=(-80,-80,-120,5,5,-15,-15,-15)$ and
$\mathbf{u}=(900,900,300,128,32,15,15,15)$ are fixed normalization constants,
not clipping bounds. World-space memory, transition reference positions, and
displacement composition retain FP32 physical coordinates.

\subsection{Joint Geometry and History Encoding}
\label{sec:layout-encoding}

\textbf{Volumetric geometry.}
The geometry encoder receives $K_t^-\odot V_t^-$, $K_t^-$, and voxel-center
height relative to the player divided by 100. Residual blocks combine
horizontal $1\times3\times3$ and vertical $3\times1\times1$ convolutions,
GroupNorm, and GELU. Table~\ref{tab:layout-architecture} specifies the feature
pyramid and Transformer sizes. The coarsest volume is projected to width 512
and flattened into 500 tokens with 3D sinusoidal position encodings. Finer
levels support reconstruction and entity-local geometry conditioning.

\begin{table}[!htb]
    \centering
    \caption{\textbf{World-state architecture.} Volume dimensions follow
    $Z,Y,X$ order. Main Transformer blocks use width 512 and feed-forward
    width 2048; entity-local attention uses width 128.}
    \label{tab:layout-architecture}
    \small
    \begin{minipage}[t]{0.46\linewidth}
        \centering
        \begin{tabular}{@{}lcr@{}}
            \toprule
            Level & Resolution & Channels \\
            \midrule
            Encoder 1 & $40\times160\times160$ & 8 \\
            Encoder 2 & $40\times80\times80$ & 16 \\
            Encoder 3 & $20\times40\times40$ & 32 \\
            Encoder 4 & $10\times20\times20$ & 64 \\
            Encoder 5 & $5\times10\times10$ & 128 \\
            \bottomrule
        \end{tabular}
    \end{minipage}\hfill
    \begin{minipage}[t]{0.51\linewidth}
        \centering
        \begin{tabular}{@{}lrr@{}}
            \toprule
            Attention module & Layers & Heads \\
            \midrule
            Shared Context Transformer & 4 & 8 \\
            Discovery Transformer & 2 & 8 \\
            New-entity Decoder & 2 & 8 \\
            Transition Transformer & 2 & 8 \\
            Entity-local attention & 1 & 4 \\
            \bottomrule
        \end{tabular}
    \end{minipage}
\end{table}

\textbf{Player and object histories.}
Player tokens combine normalized state, orientation, and an action embedding;
object tokens combine normalized state and a class embedding. For incomplete
histories, adjacent-frame position differences carry identity-aware validity:
\begin{equation}
    b_\tau^i = \mathbf{1}[\text{same object valid at }\tau-1,\tau],
    \qquad
    \mathbf{d}_\tau^i=b_\tau^i(\mathbf{p}_\tau^i-\mathbf{p}_{\tau-1}^i).
    \label{eq:layout-history-delta}
\end{equation}
A learned projection of $[\mathbf{d}_\tau^i/10,b_\tau^i]$ is added to the
corresponding object token. A missing pair has $b=0$, whereas an observed
stationary pair has $b=1$ and $\mathbf{d}=0$. Single-frame histories therefore
remain distinguishable from observed stationary histories.

Temporal, slot, and token-type embeddings identify history tokens. The Shared
Context Transformer jointly encodes geometry, player history, and masked object
history into $\mathcal{C}_t$, excluding current-frame entrants and future states.

\subsection{Scene Discovery at a Committed Pose}
\label{sec:layout-discovery}

Discovery completes the scene at the committed player pose at $t$; transition
then advances its entities to $t+1$. The crop origin thus remains independent
of the same-step predicted pose. The Discovery Transformer processes Geometry
Queries and Spatial Entity Queries jointly, with cross-attention to
$\mathcal{C}_t$.

\textbf{Geometry completion.}
The Geometry Decoder reshapes the 500 Geometry Query outputs into a coarse
volume and reconstructs occupancy using trilinear upsampling and encoder skip
connections. Separate voxel heads predict occupancy and an auxiliary
new-entity heatmap. Only unavailable voxels receive occupancy updates. The
heatmap provides training supervision, not query selection.

\textbf{Spatial entity proposals.}
Training and inference share deterministic anchor rules
(Table~\ref{tab:layout-proposals}), using only available state. After clipping
and deduplication, anchors occupy up to 128 query slots; unused slots are
masked. For player displacement exceeding 0.5 units on any axis, 44 boundary
samples favor motion-leading faces and 20 retain six-face safety coverage.
Otherwise all 64 use six-face coverage, with horizontal faces weighted twice
as strongly as top and bottom. Each Spatial Entity Query combines a learned
embedding with anchor, family, and applicable source-slot embeddings.

\begin{table}[!htb]
    \centering
    \caption{\textbf{Candidate construction.} Budgets precede clipping and
    deduplication. Matching uses a 100-unit Euclidean radius. The default
    objectness and occupancy probability thresholds are both 0.5.}
    \label{tab:layout-proposals}
    \small
    \begin{tabular*}{\linewidth}{@{\extracolsep{\fill}}llr@{}}
        \toprule
        Family & Sampling pattern & Maximum count \\
        \midrule
        Boundary & Motion-leading faces and six-face safety & 64 \\
        Active entity & Base, center, top, and nearby offsets & 32 \\
        Viewing ray & 4 depths from 40 to 400; height offsets $-20,0,20$ & 12 \\
        Interior & $3\times3\times2$ grid within the crop & 18 \\
        \bottomrule
    \end{tabular*}
\end{table}

After the Discovery Transformer, Spatial Entity Queries attend to local features
from the encoder's $40\times80\times80$ level and to a global memory of
$10\times20\times20$ encoder features through the New-entity Decoder.
They predict objectness, foreground class, and continuous attributes. New-entity
positions are bounded residuals relative to their anchors:
\begin{equation}
    \hat{\mathbf{p}}_t^j
    =\mathbf{a}_t^j+100\tanh(\mathbf{u}_t^j).
    \label{eq:layout-new-position}
\end{equation}
Height and radius use scaled sigmoid outputs, and velocity uses scaled
$\tanh$ outputs. Objectness is predicted separately from foreground identity,
separating existence prediction from semantic classification.

Supervision uses a coordinate-only Hungarian assignment restricted to reachable
anchor--target pairs, maximizing the number of distinct matched objects before
minimizing normalized position cost. Reachability requires distance at most
100 world units and per-axis offsets within the output range. Additional nearby
anchors regress to their nearest reachable target, while primary assignments
preserve distinct target coverage; unreachable targets remain unmatched.
At inference, proposals above the objectness threshold are ranked by objectness
times foreground-class probability. Class-aware non-maximum suppression treats
two proposals as duplicates when their vertical intervals overlap and their
horizontal distance is at most $\max(5,(r_i+r_j)/2)$. Filtering against existing
same-class entities additionally requires positive vertical overlap of at least
half the smaller height. Accepted entities enter free slots with one valid
history frame.

\subsection{Geometry-Conditioned State Transitions}
\label{sec:layout-transition}

The Transition Transformer attends to shared context concatenated with Current
Entity Tokens, including newly committed entities, while reusing the original
encoder features. Each non-player Entity Query is conditioned through an MLP
on its current class and state, three history differences, and their validity
bits.

To supply finer geometry, we trilinearly sample a $5\times5\times5$ neighborhood
from the encoder's $40\times80\times80$ feature volume. For persistent objects,
the samples span $\pm20$ world units horizontally and extend from ten units
below the object's base to ten units above its top. Relative sample positions
are embedded, and entity-local attention produces an additive query condition:
\begin{equation}
    \begin{aligned}
        \bar{q}_t^i &= q^i+e_{\mathrm{slot}}^i
            +F_{\mathrm{state}}(o_t^i,\mathbf{d}_{t-2:t}^i,b_{t-2:t}^i),\\
        q_{t,\mathrm{cond}}^i &= \bar{q}_t^i
            +F_{\mathrm{local}}(\bar{q}_t^i,\mathcal{P}_t^i),
    \end{aligned}
    \label{eq:layout-query-conditioning}
\end{equation}
where $\mathcal{P}_t^i$ contains sampled geometry features. Conditioning is
applied once before the Transition Transformer. Both the new-entity and
persistent-entity geometry paths backpropagate into the encoder; neither reads
newly decoded occupancy. The player query accesses geometry and action history
through the shared context.

Separate heads predict object XYZ displacement, height/radius, velocity, and
class; the player additionally predicts a normalized orientation vector.
Positions follow Equation~\ref{eq:world-displacement}: the learned displacement
is added to the current position, with no prescribed velocity integration.
A background class permits object retirement. State transitions are
deterministic given their inputs.

\subsection{Training Objectives and History Augmentation}
\label{sec:layout-training}

We train on reset-safe five-frame windows: four context frames and one
next-state target. Static targets and newly discovered objects belong to the
last context frame $t$, whereas player and persistent-object targets belong to
$t+1$. The static input is the preceding crop reprojected to the origin at $t$;
its overlap defines the known mask. Current-frame entrants are hidden from
discovery and teacher-forced into the transition input with a single valid
frame. Instance identities define persistent supervision, and crop exits are
distinguished from actual disappearances. Training uses one-step supervision.

\textbf{Static geometry.}
Unknown-voxel completion combines focal and volumetric Dice losses, with a
smaller reconstruction term on known voxels:
\begin{equation}
    \mathcal{L}_{\mathrm{static}}
    =10\mathcal{L}_{\mathrm{focal}}^{\mathrm{unknown}}
     +\mathcal{L}_{\mathrm{Dice}}^{\mathrm{unknown}}
     +0.1\mathcal{L}_{\mathrm{BCE}}^{\mathrm{known}}.
    \label{eq:layout-static-loss}
\end{equation}
Known and unknown terms have independent denominators. Dice is computed over
each full 3D crop rather than independently per height slice, excluding crops
with no unknown voxels. The focal loss uses $\alpha=0.25$ and $\gamma=2$.
A Gaussian focal objective supervises the auxiliary new-object heatmap.

\textbf{Player and persistent entities.}
The player objective is $\mathcal{L}_{\mathrm{ego}}=
5\mathcal{L}_{\mathrm{ego,geom}}+\mathcal{L}_{\mathrm{ego,vel}}$, using mean L1
errors on $(x,y,z,h,r)$ and velocity, respectively. Orientation has a separate
L1 term $\mathcal{L}_{\mathrm{angle}}$ on $(\sin\theta,\cos\theta)$.
Persistent entities use focal classification ($\gamma=2$) and L1 attribute
regression, with separate classification and regression counts. Attribute
errors sum over channels and average over valid entities. Their objective is
\begin{equation}
    \mathcal{L}_{\mathrm{pers}}=\mathcal{L}_{\mathrm{pers,cls}}
      +5\mathcal{L}_{\mathrm{pers,hr}}+\mathcal{L}_{\mathrm{pers,vel}}
      +\mathcal{L}_{\mathrm{motion}}+\mathcal{L}_{\mathrm{item}}.
    \label{eq:layout-persistent-loss}
\end{equation}
Non-item XYZ supervision uses the six phases in
Table~\ref{tab:layout-motion-phases}. Within each phase, we average summed
per-axis Smooth L1 errors between predicted and target displacements, both
divided by 10, with $\beta=0.1$.
Let $\bar\ell_{\mathrm{move}}$ average the nonempty continue, start, and
unknown-moving phase means, and let $\bar\ell_{\mathrm{stay}}$ average the
nonempty stay and unknown-stationary means. Then
\begin{equation}
    \mathcal{L}_{\mathrm{motion}}
      =5\bar\ell_{\mathrm{move}}
       +\bar\ell_{\mathrm{stay}}+2\ell_{\mathrm{stop}}.
    \label{eq:layout-motion-loss}
\end{equation}
Empty groups contribute zero; phase counts are aggregated across devices.
Separately normalized phases balance motion against frequent stationary
examples. Stationary interactable item classes are excluded from these groups;
$\mathcal{L}_{\mathrm{item}}$ averages their summed XYZ L1 errors divided by 10.

\begin{table}[!htb]
    \centering
    \caption{\textbf{Motion phases.} Moving means a 3D displacement norm
    greater than one world unit. Previous motion uses clean history with the
    post-truncation validity mask; target motion uses clean $t\to t+1$ states.}
    \label{tab:layout-motion-phases}
    \small
    \begin{tabular*}{\linewidth}{@{\extracolsep{\fill}}llll@{}}
        \toprule
        Phase & Previous pair valid & Previous motion & Target motion \\
        \midrule
        Continue & Yes & Moving & Moving \\
        Start & Yes & Stationary & Moving \\
        Stop & Yes & Moving & Stationary \\
        Stay & Yes & Stationary & Stationary \\
        Unknown-moving & No & Unavailable & Moving \\
        Unknown-stationary & No & Unavailable & Stationary \\
        \bottomrule
    \end{tabular*}
\end{table}

\textbf{New entities and joint objective.}
New entities use sigmoid focal objectness ($\alpha=0.25$, $\gamma=2$),
foreground focal classification ($\gamma=2$) on positive anchors, and Smooth L1
regression on normalized attributes, weighted $5:1:1$ for XYZ, height/radius,
and velocity. Objectness weight increases linearly from 0.1 to 1 over the first
500 training steps. These terms form $\mathcal{L}_{\mathrm{new}}$ and share a
normalizer equal to the global positive-query count divided by device count,
or one if there are no positives. GIoU is monitored only. The joint objective is
\begin{equation}
    \begin{aligned}
    \mathcal{L}_{\mathrm{state}}
      ={}&5\mathcal{L}_{\mathrm{static}}+10\mathcal{L}_{\mathrm{heat}}
          +20\mathcal{L}_{\mathrm{ego}}+10\mathcal{L}_{\mathrm{angle}}\\
         &+\mathcal{L}_{\mathrm{pers}}+\mathcal{L}_{\mathrm{new}}.
    \end{aligned}
    \label{eq:layout-objective}
\end{equation}

\textbf{Robustness to imperfect histories.}
We randomly permute object slots and truncate history prefixes with probability
0.3 per object, preserving the latest valid frame and allowing single-frame
histories. Player and non-item positions receive independent uniform noise
with horizontal bounds $\pm\rho r$ and vertical bounds $\pm\rho h$, using
$\rho=0.002$ and $0.02$, respectively. For stationary items, 30\% of sequences
instead receive temporally correlated drift after history truncation. For
$m\geq2$ surviving observations, the $k$th valid position is perturbed by
\begin{equation}
    \tilde{\mathbf{p}}_k^i
      =\mathbf{p}_k^i+\frac{k}{m-1}\boldsymbol{\eta}^i,
    \qquad k=0,\ldots,m-1,
    \label{eq:layout-item-noise}
\end{equation}
where horizontal components of $\boldsymbol{\eta}^i$ are uniform in $[-1,1]$
and its vertical component is uniform in $[-2,2]$ world units. The first
surviving observation stays clean, and one-frame histories are not corrupted.
Targets remain clean; conditioning differences are recomputed from the
perturbed histories, while motion-phase labels use clean differences with the
updated validity masks. Static inputs remain unperturbed.

\subsection{Rollout and Memory Management}
\label{sec:layout-rollout}
\label{app:layout-procedure}

Given the current committed player pose, retained histories, and action $a_t$,
one simulator step proceeds as follows:
\begin{enumerate}[itemsep=2pt,topsep=4pt]
    \item Crop the global occupancy and availability caches at the current pose.
    Express the retained object and player histories in this crop's coordinate
    frame and construct state-derived candidate anchors.
    \item Encode the pre-discovery histories and crop once. Retain the shared
    context tokens and the multiscale geometry features.
    \item Run the Discovery Transformer and New-entity Decoder. Run the Geometry
    Decoder when the crop contains unavailable voxels.
    \item Commit geometry at the current crop origin, preserving known voxels.
    Restore eligible dormant entities, then threshold and deduplicate discovered
    objects. Allocate clean slots and insert each accepted object into the
    current history frame.
    \item Concatenate Current Entity Tokens with the retained context and run
    the Transition Transformer without re-encoding completed occupancy.
    \item Convert predictions to world coordinates. Retire background-class
    entities, store out-of-volume entities as dormant, and append the next
    player and active-object states to history.
    \item Project available geometry and updated entities into depth and
    semantic conditions at the new viewpoint. Generated RGB updates only visual
    history; the next simulator step completes newly exposed map regions.
\end{enumerate}

The map cache preserves both occupied and empty committed voxels on revisits.
Dormant entities retain their last state outside the local volume and resume
updates on re-entry; reallocated slots have their histories cleared. Skipping
the Geometry Decoder leaves both entity branches active because they obtain
spatial features from the encoder.
\FloatBarrier

\section{Renderer Details}
\label{app:renderer}

This appendix specifies world-state projection, renderer architecture,
training stages, and cached inference described in Section~\ref{sec:renderer}.

\subsection{World-State Projection}
\label{sec:renderer-projection}

The deterministic operator $\Pi$ converts occupancy and entity states into
aligned $160\times120$ semantic and depth conditions. Camera rays use the
player's position and yaw, the configured pitch, and a $90^\circ$ horizontal
field of view, following Doom's $320/200$ projection aspect ratio. Rays march
through the local occupancy grid in 2.5-unit increments, stopping at the first
occupied voxel, the crop boundary, or a 400-unit range limit. Surface hits
provide camera-forward depth and a shared static-geometry label (255).
Each non-player entity is represented by an axis-aligned box of size
$2r\times2r\times h$ at its base position. Box faces are clipped at a one-unit
near plane and rasterized from far to near. A face's mean camera depth is
checked against static depth at its projected polygon center before filling
the polygon with that depth and the entity's class ID. The same lightweight
projection convention is used for data generation and learned-world rollout;
surface textures are supplied by the neural renderer.

For a projected surface with camera-forward depth $D_t(u,v)$, the normalized
depth condition is
\begin{equation}
    d_t(u,v)=Q_8\!\left(1-\operatorname{clip}
        \left(\frac{D_t(u,v)}{400},0,1\right)\right),
    \qquad Q_8(q)=\frac{\lfloor255q\rfloor}{255}.
    \label{eq:renderer-depth-encoding}
\end{equation}
This encoding gives larger values to nearer surfaces and preserves the
8-bit quantization used in the training data. Pixels with no projected surface
have semantic label $s_t=0$ and depth value $d_t=0$. The renderer receives
separate validity masks $m_t^s=\mathbf{1}[s_t\ne0]$ and
$m_t^d=\mathbf{1}[d_t>0]$ alongside these maps.

\subsection{Architecture}
\label{sec:renderer-architecture}

The renderer predicts frame $x_t$ from four preceding RGB frames, target-view
conditions $c_t=(s_t,d_t,m_t^s,m_t^d)$, and action $a_{t-1}$. Here $s_t$ and
$d_t$ are semantic labels and normalized depth as defined above, with validity masks
$m_t^s,m_t^d$. A frozen, domain-trained VAE encodes each $120\times160$ image
into a $4\times30\times40$ latent using its posterior mean. Latents are
standardized channel-wise before diffusion and unstandardized before decoding.
The history and scene-condition encoders are separate from target denoising.

\textbf{RGB-history encoding.}
Each latent is divided into $2\times2$ patches, producing 300 tokens per frame
and 1,200 for a complete four-frame history. Learned spatial and relative
frame-position embeddings identify location and time. A two-block Transformer
attends over valid history tokens to produce $H_t$; all history frames precede
the target.

\textbf{Scene-condition encoding.}
Categorical semantic embeddings are concatenated with normalized depth and both
validity masks. A convolutional encoder reduces them to latent resolution;
patch projection and spatial/type embeddings produce scene-condition tokens
$C_t$. Token validity follows the input masks. Keeping $C_t$ separate from
$H_t$ gives each denoising block direct access to current geometry and semantics
alongside historical appearance.

\textbf{Target denoiser.}
The noisy target uses the same latent patch projection, with spatial and
target-type embeddings. Each of eight blocks applies target self-attention,
cross-attention to $H_t$, cross-attention to $C_t$, and a feed-forward network,
using pre-normalization and residual connections. Summed noise-index and
action embeddings are projected in each block and added before query and
feed-forward normalizations. A linear output head reconstructs the latent
grid. Both Transformers use width 384, six attention heads, feed-forward
width 1,536, and zero dropout.

\subsection{Training Stages and History Construction}
\label{sec:renderer-objective}
\label{sec:renderer-mixed-history}

\textbf{Denoising objective.}
Diffusion perturbs only the target latent; history and scene conditions are
held fixed. For standardized target $z_t$ and $\epsilon\sim\mathcal{N}(0,I)$,
\begin{equation}
    z_t^{(n)}=\alpha_n z_t+\sigma_n\epsilon,
    \qquad
    v_t^{(n)}=\alpha_n\epsilon-\sigma_n z_t,
    \label{eq:renderer-diffusion}
\end{equation}
where $\alpha_n^2+\sigma_n^2=1$. We use 1,000 diffusion indices with linearly
spaced variances from $10^{-4}$ to $0.02$. The base objective is v-prediction MSE:
\begin{equation}
    \mathcal{L}_{\mathrm{diff}}
    =\mathbb{E}\!\left[
        \|v_\theta(z_t^{(n)},H_t,C_t,a_{t-1},n)-v_t^{(n)}\|_2^2
      \right],
    \label{eq:renderer-v-loss}
\end{equation}
averaged over latent elements. Table~\ref{tab:renderer-training-stages}
distinguishes the training stages. The final 90,000 updates use this MSE with
uniformly sampled diffusion indices and a mixture of 50\% ground-truth history,
40\% generated history, and 10\% history dropout. Ground-truth examples supply
the four recorded frames preceding the target. The VAE stays frozen throughout.

\begin{table}[!htb]
    \centering
    \caption{\textbf{Renderer training stages.} Intermediate refinement adds
    clean-latent value and spatial-gradient L1 terms for $n\leq200$, averaged
    over eligible samples. Gradient error averages horizontal and vertical
    finite differences. Both auxiliary terms are disabled for the final stage.}
    \label{tab:renderer-training-stages}
    \small
    \setlength{\tabcolsep}{4pt}
    \begin{tabular}{@{}>{\raggedright\arraybackslash}p{0.17\linewidth}
                    >{\raggedright\arraybackslash}p{0.32\linewidth}
                    >{\raggedright\arraybackslash}p{0.45\linewidth}@{}}
        \toprule
        Stage & RGB history & Objective \\
        \midrule
        Base training & Ground-truth & $\mathcal{L}_{\mathrm{diff}}$ \\
        Refinement & Ground-truth
        & $\mathcal{L}_{\mathrm{diff}}$ + value L1 (0.25) + gradient L1 (0.1) \\
        Mixed-history & 50\% ground-truth, 40\% generated, 10\% none
        & $\mathcal{L}_{\mathrm{diff}}$ \\
        \bottomrule
    \end{tabular}
\end{table}

\textbf{Generated RGB history.}
We cache short autoregressive rollouts from the frozen step-102,500 renderer.
Each anchor contains four recorded initial frames followed by four generated
frames, using recorded scene conditions and actions. We uniformly sample a
depth $k\in\{1,2,3,4\}$ and take the last four frames after $k$ generated
frames as history. The target is the \emph{next recorded frame}, with its
aligned scene conditions and preceding action. Thus histories contain one to
four generated frames while targets stay clean. Anchors use non-overlapping
nine-frame windows within reset-safe trajectory segments. The offline cache
remains fixed during fine-tuning, so each update requires one target-denoising
evaluation per example rather than an online rollout.

\textbf{RGB-history dropout.}
All RGB-history slots are marked invalid while scene conditions and action
remain available. The history encoder supplies a content-free fallback. This
trains condition-only synthesis and supports initialization without reference
RGB frames.

\subsection{Efficient Sampling}
\label{sec:renderer-inference}

Each frame starts from Gaussian latent noise and uses eight denoising steps
by default, with indices evenly spaced from 999 to 0 and rounded to integers.
Updates combine the predicted clean latent and noise at the next index without
injecting additional noise. The final clean prediction is unstandardized,
decoded to RGB, and retained directly as a latent history entry, replacing the
oldest frame without re-encoding the generated image.

History and scene conditions are encoded once per frame. Their tokens and
per-block cross-attention keys and values are cached across all denoising
steps; only target-dependent computation is repeated. The cache is rebuilt
for the next frame, keeping the history length and per-frame computation
independent of total rollout duration.
\FloatBarrier

\section{Evaluation Details}
\label{app:experiments}

World-state and renderer checkpoints are at steps 200,000 and 192,500.
Unseen levels are Scythe MAP01--05, MAP07, and MAP10; engine references
use ViZDoom 1.3.0 and $90^\circ$ horizontal field of view.
Other shared settings are given in Section~\ref{sec:exp-setup}.

\subsection{Interaction and Control}
\label{app:exp-state}

\textbf{Contact counting.}
CER is $100N_{\mathrm{failed}}/N_{\mathrm{resolved}}$, counting successful
and failed obstacle contacts and landings up to each horizon. Overlapping
categories count once; unfinished or unassessable encounters are excluded.
Counts persist across respawns, excluding teleportation. Contact tests use
supplied geometry in Global and generated memory in Local.

\textbf{Monster pursuit.}
\label{app:exp-monster-target}
Trials reset one four-frame Zombieman history in the full map, remove
other entities, and disable discovery. We sample 20 stationary player
locations on a 20-unit grid, five per quadrant, 100--220 units from the
monster (seed 1234), requiring support, empty body volume, and a sampled
supported straight path. Success requires body-surface gap $\leq30$,
feet-height difference $\leq10$, and vertical overlap $\geq28$ units,
using initial body dimensions. All trials count toward the 200-step
statistic. Figure~\ref{fig:exp-monster-motion}A--D use 300 steps;
E uses a separate wall-side initialization.

\textbf{Multiplayer and interventions.}
\label{app:exp-extensions}
Examples use full-map initialization with other entities removed.
Players exchange states, representing peers as Zombiemen, with independent
actions and RGB histories and no collision correction.
Opponent-presence scores follow the VLM protocol of \citet{po2026multigen},
with visible as positive; MultiGen values are quoted from that paper.
Programmed health loss and death modify explicit state with fixed weights;
health feedback is overlaid on rendered observations.

\subsection{Visual Quality}
\label{app:exp-rendering}

\textbf{Frame and distribution metrics.}
PSNR, SSIM, and VGG LPIPS v0.1 use full $160\times120$ frames, including
weapons, averaged within clips and equally across episodes.
SSIM uses an $11\times11$ Gaussian window ($\sigma=1.5$), population
covariance, and valid centers. FID uses Inception-v3 pool3 (2048-D);
FVD uses TF-Hub Kinetics-400 I3D pre-softmax logits (400-D).
Both sides use lossless RGB, resized identically to $299\times299$ for
FID and $224\times224$ for FVD. Each 32-frame clip supplies eight FID
frames at offsets $0,4,\ldots,28$ and one FVD sample. Distribution scores
pool samples per setting and horizon, using FP32 features and FP64
statistics (sample covariance for FID; population covariance for FVD).

\textbf{References and recovery.}
\label{app:exp-local-quality}
Global frame metrics use pose-aligned engine references with native map
textures. Local FID/FVD share an unpaired Freedoom2 reference pool.
Recovery retains world memory but resets RGB history. Clips overlapping
failure, recovery, or 32 post-recovery warmup steps are excluded from
visual scores, but failures remain in CER.
PlayGen receives actions rather than explicit states; its paired scores
include trajectory divergence as well as appearance differences.

\subsection{Interactive Efficiency}
\label{app:exp-efficiency}

Both models share one M4 Pro device (MPS FP32), with one process and one
Torch CPU thread per model.
Pipelining overlaps world update $t+1$ with rendering $t$.
Table~\ref{tab:exp-efficiency} uses two 60-frame repeats,
after 12 warmup frames and reset,
with matched actions and noise. Throughput includes pipeline fill/drain
and inter-process transport, excluding initialization, network, and display.

\begin{table}[!htb]
    \centering
    \caption{\textbf{M4 Pro backend efficiency.} Steps denotes
    renderer denoising steps. Times are mean milliseconds; latency spans
    world-update start to completed RGB encoding. World includes projection
    and state handling; Render includes VAE decoding and output encoding.}
    \label{tab:exp-efficiency}
    \small
    \setlength{\tabcolsep}{4pt}
    \begin{tabular*}{\linewidth}{@{\extracolsep{\fill}}lrrrrr@{}}
        \toprule
        Schedule & Steps & FPS & Latency & World & Render \\
        \midrule
        Serial   & 4 & 6.19 & 161.2 & 82.8 & 77.9 \\
        Pipeline & 4 & 8.97 & 212.5 & 104.7 & 102.3 \\
        Serial   & 8 & 4.98 & 200.6 & 83.3 & 116.7 \\
        Pipeline & 8 & 7.16 & 275.7 & 91.7 & 137.8 \\
        \bottomrule
    \end{tabular*}
\end{table}

\FloatBarrier

\clearpage
\section{Design Comparisons}
\label{app:ablations}

We examine local geometry conditioning, volumetric map modeling, and
renderer training. Comparisons use trained checkpoints with differing
training budgets and settings rather than budget-matched retraining.

\subsection{Local Geometry Conditioning}
\label{app:ablation-local-map}

We compare 2D world-state models trained without and with encoder-side
local-map conditioning, retaining each model's trained inputs.
Both use the same 28 player placements from three NPC histories:
16 obstructed and twelve clear initial directions. Each rollout keeps
one Zombieman and a stationary player in the full reference map for at
most 300 steps, without collision correction or ground-truth NPC feedback.
Rollouts end at severe penetration, target proximity, entity loss, or
timeout. Penetration requires more than half of 125 torso samples to
overlap occupied voxels; proximity uses a 30-unit body-surface gap with
vertical compatibility. Contact counting follows Appendix~\ref{app:exp-state}.

\begin{table}[!htb]
    \centering
    \caption{\textbf{Local-map conditioning on challenging NPC placements.}
    Penetration and proximity count episodes; CER counts failed/resolved
    horizontal encounters.}
    \label{tab:ablation-local-map}
    \small
    \setlength{\tabcolsep}{5pt}
    \begin{tabular}{lccc}
        \toprule
        Model & Penetration $\downarrow$ & CER $\downarrow$
        & Proximity $\uparrow$ \\
        \midrule
        Without local map & 24/28 & 24/27 (88.9\%) & 4/28 \\
        With local map & \textbf{12/28} & \textbf{12/22 (54.5\%)}
        & \textbf{15/28} \\
        \bottomrule
    \end{tabular}
\end{table}

\begin{figure}[!htb]
    \centering
    \includegraphics[width=\linewidth]{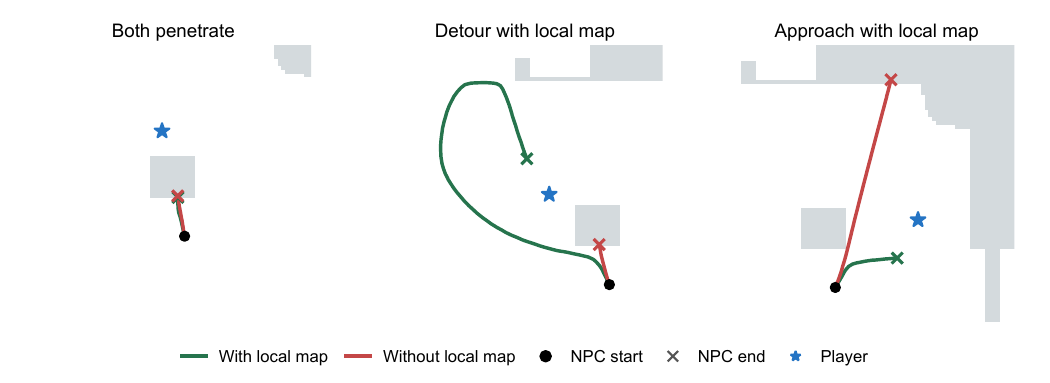}
    \caption{\textbf{Local-map conditioning and NPC trajectories.}
    Paired predictions from identical initial states, with matched spatial
    scales. Both models fail in the first example; local conditioning
    improves approach in the other two.}
    \label{fig:ablation-local-map}
\end{figure}

The local-map model halves penetration episodes and increases successful
approaches from 4/28 to 15/28 (Table~\ref{tab:ablation-local-map}).
Both variants move by more than one unit in approximately 77\% of audited
steps, so the improvement is not explained by stationary predictions.
Figure~\ref{fig:ablation-local-map} shows paired trajectories, including a
remaining failure.

\FloatBarrier
\subsection{Geometry Representation}
\label{app:ablation-geometry}

We compare the 2D encoder-local-map model with the 3D model at step
200,000 along eight fixed Freedoom2 engine trajectories, each lasting
700 transitions (60 seconds). Poses, actions, and initial local geometry
are identical; predicted map updates are committed to memory and fed
back without future reference geometry. Dynamic entities and learned
player updates are excluded to isolate map generation from trajectory drift.

\begin{table}[!htb]
    \centering
    \caption{\textbf{Geometry quality after 60-second rollouts.}
    Scores pool generated reference voxels over eight paths. Fragment density
    counts isolated six-connected components of at most 27 voxels per
    million generated voxels, excluding components touching initialization
    or unknown/crop boundaries.}
    \label{tab:ablation-geometry}
    \small
    \setlength{\tabcolsep}{10pt}
    \begin{tabular}{lrr}
        \toprule
        Metric & 2D & 3D \\
        \midrule
        Occupancy IoU (\%) $\uparrow$ & 61.66 & \textbf{64.30} \\
        Precision (\%) $\uparrow$ & \textbf{92.32} & 89.57 \\
        Recall (\%) $\uparrow$ & 64.99 & \textbf{69.50} \\
        Fragment density $\downarrow$ & 48.61 & \textbf{9.94} \\
        \bottomrule
    \end{tabular}
\end{table}

\begin{figure}[!htb]
    \centering
    \includegraphics[width=\linewidth]{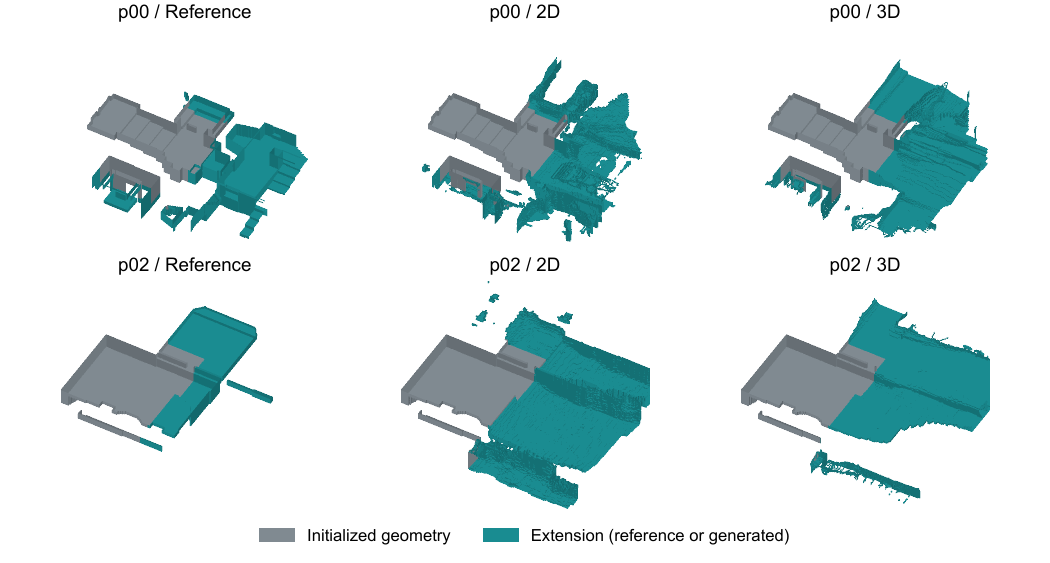}
    \caption{\textbf{Geometry after continuous generation.}
    Two matched trajectories, with identical framing within each row.
    Gray denotes initialized geometry; teal denotes reference or generated
    extensions.}
    \label{fig:ablation-geometry}
\end{figure}

The 3D model produces 79.5\% fewer small isolated components (919 versus
188), with reductions on all eight paths and the same trend under
alternative connectivity and size thresholds. Aggregate IoU and recall
increase, with a modest decrease in precision
(Table~\ref{tab:ablation-geometry}). Figure~\ref{fig:ablation-geometry}
compares generated geometry along matched paths.

\FloatBarrier
\subsection{Renderer Training}
\label{app:ablation-renderer}

We report one-minute rendering quality on Freedoom2 MAP01 with
condition-only initialization: no real RGB history is supplied, and
subsequent RGB histories are generated autoregressively.
The earlier checkpoint (step 102,500) is evaluated using fixed engine
conditions, eight denoising steps, BF16, and full-frame metrics at
$160\times120$. The final checkpoint (step 192,500) follows mixed-history
training (Appendix~\ref{sec:renderer-objective}).
Table~\ref{tab:ablation-renderer} reproduces its Seen/Global scores from
Table~\ref{tab:exp-main}; the two rows come from separate evaluation runs.

\begin{table}[!htb]
    \centering
    \caption{\textbf{Reported renderer quality with condition-only startup.}
    Seen map at one simulated minute. After reproduces the main-evaluation
    result; Before is from a separate checkpoint evaluation.}
    \label{tab:ablation-renderer}
    \small
    \setlength{\tabcolsep}{8pt}
    \begin{tabular}{lrrr}
        \toprule
        Training stage & PSNR $\uparrow$ & SSIM $\uparrow$ & LPIPS $\downarrow$ \\
        \midrule
        Before & 17.1 & 0.31 & 0.62 \\
        After (Table~\ref{tab:exp-main}) & 18.3 & 0.40 & 0.41 \\
        \bottomrule
    \end{tabular}
\end{table}
\FloatBarrier

\end{document}